\documentclass[10pt,twocolumn,letterpaper]{article}

\usepackage{wacv2023/wacv}
\usepackage{times}
\usepackage{epsfig}
\usepackage{graphicx}
\usepackage{amsmath}
\usepackage{amssymb}
\usepackage{booktabs}
\usepackage[accsupp]{axessibility}

\usepackage{multirow}

\def\wacvPaperID{1138} 

\wacvalgorithmstrack   

\wacvfinalcopy 

\usepackage[capitalize]{cleveref}
\crefname{section}{Sec.}{Secs.}
\Crefname{section}{Section}{Sections}
\Crefname{table}{Table}{Tables}
\crefname{table}{Tab.}{Tabs.}

\makeatletter
\def\thanks#1{\protected@xdef\@thanks{\@thanks
        \protect\footnotetext{#1}}}
\makeatother
 
\begin{document}

\title{MonoEdge: Monocular 3D Object Detection Using Local Perspectives}

\author{Minghan Zhu\textsuperscript{1*}, 
Lingting Ge\textsuperscript{2}, 
Panqu Wang\textsuperscript{2}, 
Huei Peng\textsuperscript{1}\\
\textsuperscript{1}University of Michigan\\
\textsuperscript{2}TuSimple Inc\\
{\tt\small minghanz@umich.edu, lingting.ge@tusimple.ai, panqu.wang@tusimple.ai, hpeng@umich.edu}
\thanks{*Work done during internship at TuSimple Inc. }
}

\maketitle
\thispagestyle{empty}

\begin{abstract}
   We propose a novel approach for monocular 3D object detection by leveraging local perspective effects of each object. While the global perspective effect shown as size and position variations has been exploited for monocular 3D detection extensively, the local perspectives has long been overlooked. 
   We design a local perspective module to regress a newly defined variable named keyedge-ratios as the parameterization of the local shape distortion to account for the local perspective, and derive the object depth and yaw angle from it. Theoretically, this module does not rely on the pixel-wise size or position in the image of the objects, therefore independent of the camera intrinsic parameters. By plugging this module in existing monocular 3D object detection frameworks, we incorporate the local perspective distortion with global perspective effect for monocular 3D reasoning, and we demonstrate the effectiveness and superior performance over strong baseline methods in multiple datasets.
\end{abstract}

\section{Introduction}
\label{sec:intro}

3D object detection is an important perception task for autonomous driving and other robotic applications. Monocular approaches are considered challenging mainly due to the lost of depth information in a single image. However, monocular 3D object detection still attracts a lot of interest, partly because of the low cost compared with LiDAR and the simple sensor setup compared with stereo cameras. Despite the absence of depth in a single image, monocular 3D inference is possible in practice through learning, due to the implicit prior knowledge of the physical scene layout and size of objects embedded in the data. Given such prior knowledge, the 3D localization of objects is connected with their size and position in the image plane, through the camera geometry. 

\begin{figure}[t]
  \centering
   \includegraphics[width=0.9\linewidth]{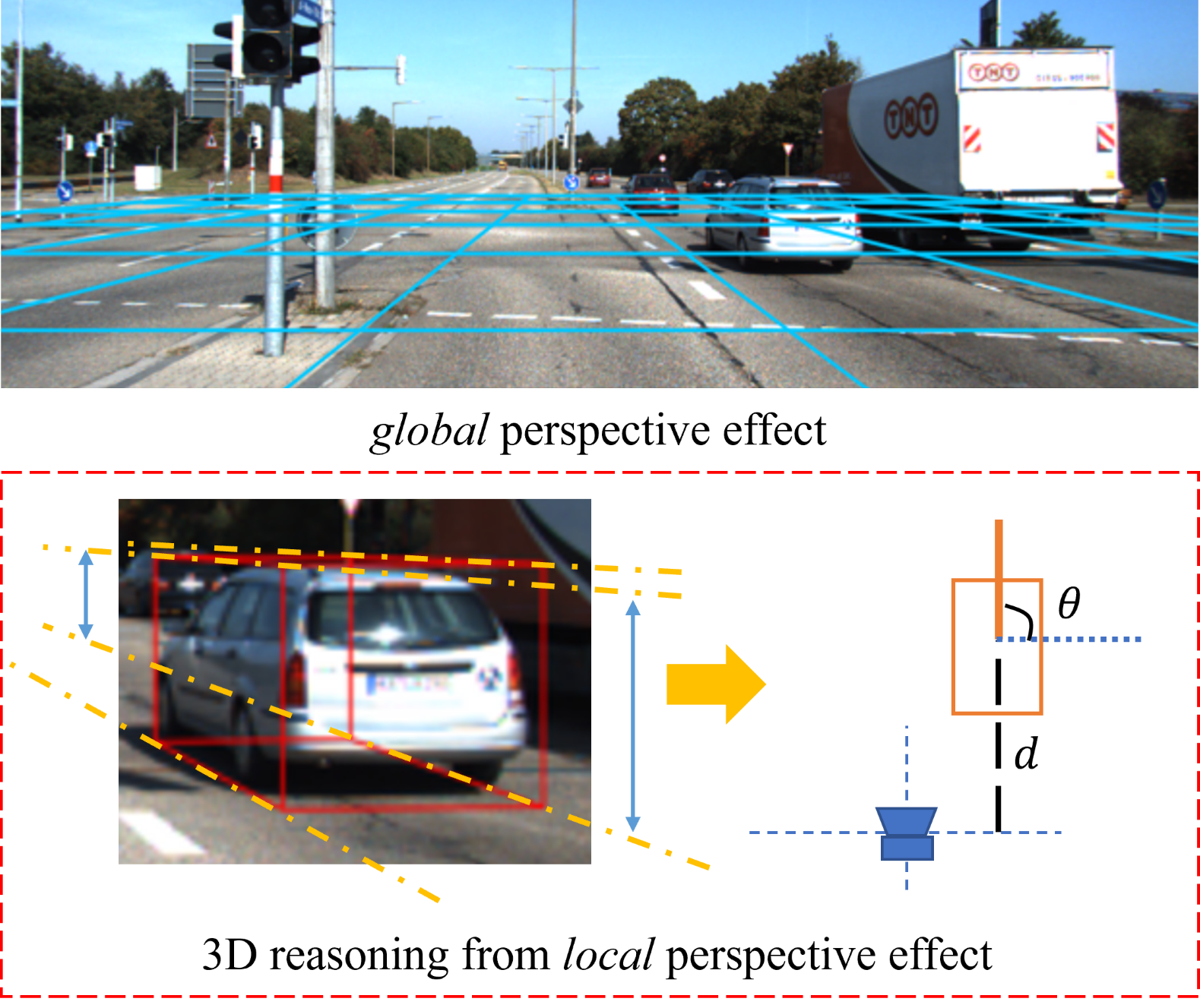}
   \caption{Up: the \textit{global} perspective effect of appearance and size change along the depth direction of an image. Down: the \textit{local} perspective effect is embedded in the shape distortion \textit{inside} an object, as highlighted by the dashed lines. We propose to infer the depth $d$ and yaw angle $\theta$ from the local perspectives. }
   \label{fig:title}
\end{figure}

Position and size are two major clues leveraged by existing monocular 3D object detection methods, and they are the result of global perspective effects on the whole image. For a camera mounted on a vehicle with viewing angle similar to the sight of human and common scenes where most objects are on the approximately flat ground, closer objects appear near the bottom of image, and the position of further objects are higher near the horizon. It is used as a prior knowledge for depth estimation in several works \cite{liu2021ground,brazil2019m3d}. In addition, objects appears larger when they are close and smaller otherwise. This relation is also often applied in the inference of depth \cite{zhang2021objects,zhang2021learning}. Meanwhile, some methods rely on pixel-wise depth image as an input or middle output \cite{manhardt2019roi}. As discussed in \cite{dijk2019neural}, a dense depth prediction network heavily relies on the pixel position in the image for depth prediction as well. 

In this work, we propose to explore another type of clues for 3D reasoning: \textit{local} perspectives. Here we aim to exploit the local distortion \textit{inside} an object, rather than among different objects. In other words, the closer part of an object appears larger than the remote part, and the magnitude of such distortion is directly connected with its distance. See \cref{fig:title} for a comparison between global perspectives and local perspectives. Capturing the perspective distortion of an object is a common skill in drawing, and it largely affects how real a painting looks, as it is tightly connected with the 3D perception of human beings. Inspired by this observation, we want to exploit this factor in improving the 3D inference of monocular computer vision. 

To achieve this goal, we first propose a method to parameterize the magnitude of local perspective distortion, by defining the \textit{keyedges} and the \textit{keyedge-ratios}. Then we show that both the depth and yaw angle of an object can be determined given the keyedge-ratios and the physical size. Our orientation estimation is also novel in that the global yaw angle is directly obtained, rather than following most existing approaches that estimate the allocentric observation angle first \cite{mousavian20173d}. Our approach, named MonoEdge, can be formed as a lightweighted module in the neural networks, and can be plugged in to existing monocular 3D detection frameworks. Our experiments demonstrate effectiveness and superior performance over strong baseline methods on the KITTI 3D detection benchmark \cite{Geiger2012CVPR}, and the improvement is also validated on the nuScenes dataset \cite{nuscenes2019}. 

In summary, our work has the following contributions: 
\begin{itemize}
    \item We propose to leverage local perspective distortion for monocular 3D object detection. 
    \item Through our approach, the depth and global yaw angle can be estimated from the local appearance of an object in an image, without knowing the camera intrinsic parameters. 
    \item We show the general applicability of our local-perspective-based method by incorporating it with strong baseline methods on the KITTI dataset \cite{Geiger2012CVPR} and on the nuScenes dataset \cite{nuscenes2019}, and improving them respectively. 
\end{itemize}

\section{Related work}

\subsection{Object and pixel depth estimation}
As mentioned in several previous work \cite{ma2021delving,wang2021probabilistic}, inaccurate depth is a dominant single factor limiting the accuracy of monocular 3D object detection. Much research effort are dedicated to improving the object based depth estimation. A simplest strategy is to directly regress the depth of objects\cite{liu2020smoke,wang2021fcos3d}, and it actually works quite well in benchmarks like KITTI \cite{Geiger2012CVPR} and nuScenes \cite{nuscenes2019}. However, the directly regressed depth is generally overfitted to the camera intrinsic and extrinsic parameters of the training data, and performs poorly when the camera parameters are changed. \cite{zhou2021monocular} partially addressed this issue by estimating and compensating the extrinsic perturbation. Pixel-wise depth regression is a similar task, which is also used as an input or middle-output in some monocular 3D detection work \cite{manhardt2019roi,ding2020learning}. It suffers from similar difficulty in generalizing to different camera parameters \cite{facil2019cam}. As revealed in \cite{dijk2019neural}, the depth regression relies heavily on the height of the pixel in the image, implying the dependency on pixel positions and camera parameters. 

The qualitative relation that pixels higher and nearer to the horizon are generally further away can serve as a prior for depth estimation. \cite{brazil2019m3d,liu2021ground} incorporated such prior into the network design by treating each row of the feature map differently in the convolution. Visual size is another important clue for 3D inference,  
and \cite{zhang2021objects,zhang2021learning} estimate the height of objects and then calculate the depth from it. \cite{shi2021geometry,lu2021geometry} focused on the uncertainty analysis of depth estimated from the appearance and physical height of objects. Overall, the position and size priors are governed by the global perspective effect on the whole image. \cite{wang2021probabilistic,chen2020monopair} model the global perspective relation among objects informed by their positions in the image. \cite{lian2021geometry} explored data augmentation with size-position relation that is compliant to the global perspectives. 

\subsection{Object orientation estimation}\label{sec:review_rot}
Orientation estimation is another important topic for monocular 3D object detection. Deep3DBox \cite{mousavian20173d} shows that the visible part of an object in an image is mainly determined by the local observation angle (also called the allocentric angle), instead of the global yaw angle (also called the egocentric angle, see \cref{sec:geometry}). Since the visible part largely determines the appearance of an object in an image, they first regress the allocentric angle from the network, and then convert it to the egocentric yaw angle with the pixel position and camera intrinsics. 
This strategy is widely adapted in later works \cite{kundu20183d,manhardt2019roi}. Recently, Ego-Net \cite{li2021exploring} directly regresses the egocentric yaw angle from a network, which proposed a progressive lifting strategy with intermediate geometrical representations. 

\subsection{Perspective effect in 3D object detection}
Some previous work contains the idea of using local shape distortion in 3D inference. GS3D \cite{li2019gs3d} extracts features from the visible faces of a coarse 3D bounding box through perspective transform. KM3D-Net \cite{li2021monocular} designs a Vanishing-Point-to-Orientation network module to regress orientation from the projected 3D box corner points in the image plane. Some previous work follows a render-and-compare strategy \cite{kundu20183d,ku2019monocular,chen2021monorun,liu2021autoshape}, which optimize the 3D properties by aligning the rendered shape model with the 2D observation. 
However, they do not explicitly parameterize the local perspective distortions or derive 3D properties from this information, as we will show in this paper. 

\section{Methodology}
We first introduce the basic geometry and the definitions of variables in  our model (\cref{sec:geometry}). We then 
derive the depth and global yaw angle using local perspectives (\cref{sec:kedge_depth}). The network design for keyedge-ratio regression is presented in \cref{sec:network}. 

\subsection{Preliminaries}\label{sec:geometry}

\begin{figure}[t]
  \centering
  \includegraphics[width=0.8\linewidth]{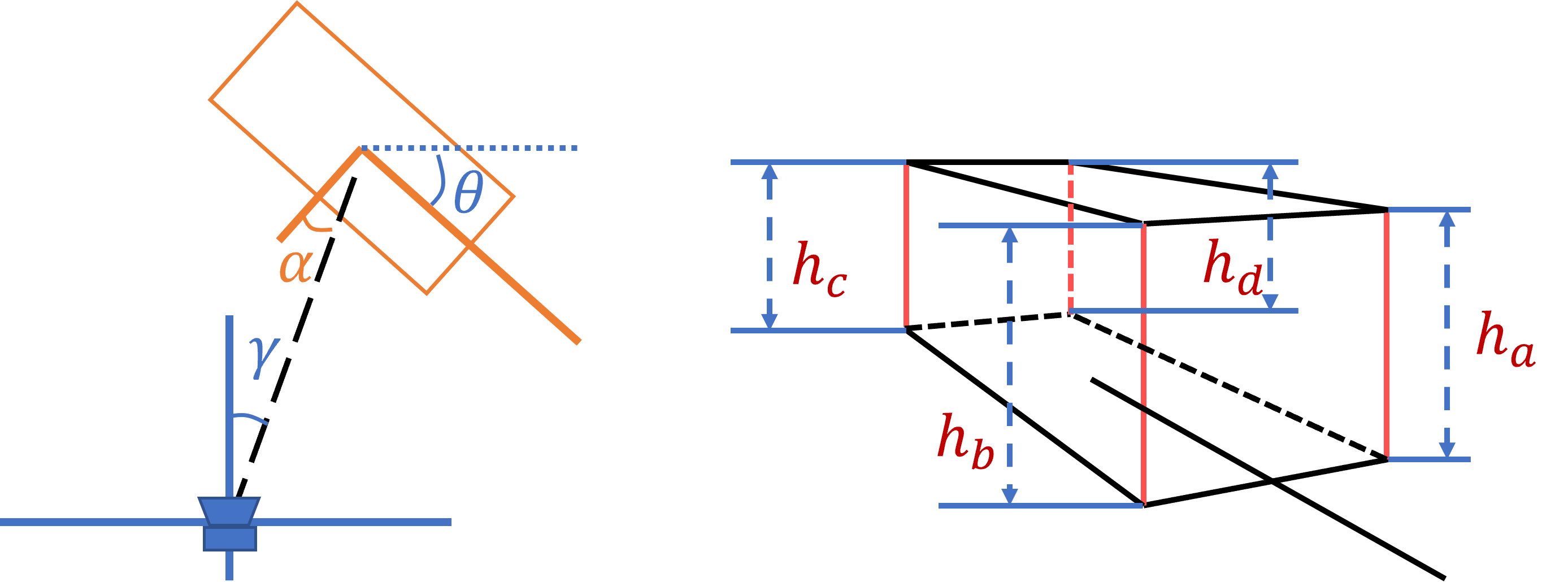}

   \caption{Left: the geometry and angle definition in 3D detection. All the angles denoted are shown in their positive direction according to KITTI's definition. Right: The projection of a 3D bounding box in an image, and the \textit{keyedges} are highlighted in red. }
   \label{fig:geometry}
\end{figure}



First we explain the definition of parameters used in our model. As shown on the left of Fig. \ref{fig:geometry}, the global yaw angle (or the \textit{egocentric angle}) of an object is the angle between the camera's right axis and the object's front axis, denoted as $\theta$. The local observation angle (or the \textit{allocentric angle}) is the angle of the camera's position w.r.t. the object's right axis, denoted as $\alpha$. The angle of the object's position w.r.t. the camera's front axis is also shown as $\gamma$, and we call it the \textit{viewing angle}. They satisfy the equation $\theta = \alpha + \gamma$. Among these angles, $\gamma$ could be calculated from the pixel's horizontal position as long as the camera intrinsic parameters are given. 

The local perspective effect is shown as distortion inside an object where the further part appears smaller. Since most objects are aligned to the vertical axis (zero pitch and roll), the size distortion can be measured conveniently through the height change. Therefore, we define the four vertical edges in a 3D bounding box as \textit{keyedges}, and use the ratio between the visual height of the keyedges to parameterize the local perspective distortion. We call them \textit{keyedge-ratios}: 
\begin{equation}
    r_{ij} \triangleq \frac{h_i}{h_j}
\end{equation}
where $h_i$ and $h_j$ are the visual height of the keyedges $i$ and $j$. The keyedges are indexed in an object-centric way: indices $a, b, c, d$ are assigned clockwisely starting from the front-left corner of the object, as shown on the right of \cref{fig:geometry}.

\subsection{Intrinsics-free depth and yaw angle derivation}\label{sec:kedge_depth}

\begin{figure}[t]
  \centering
  \includegraphics[width=0.8\linewidth]{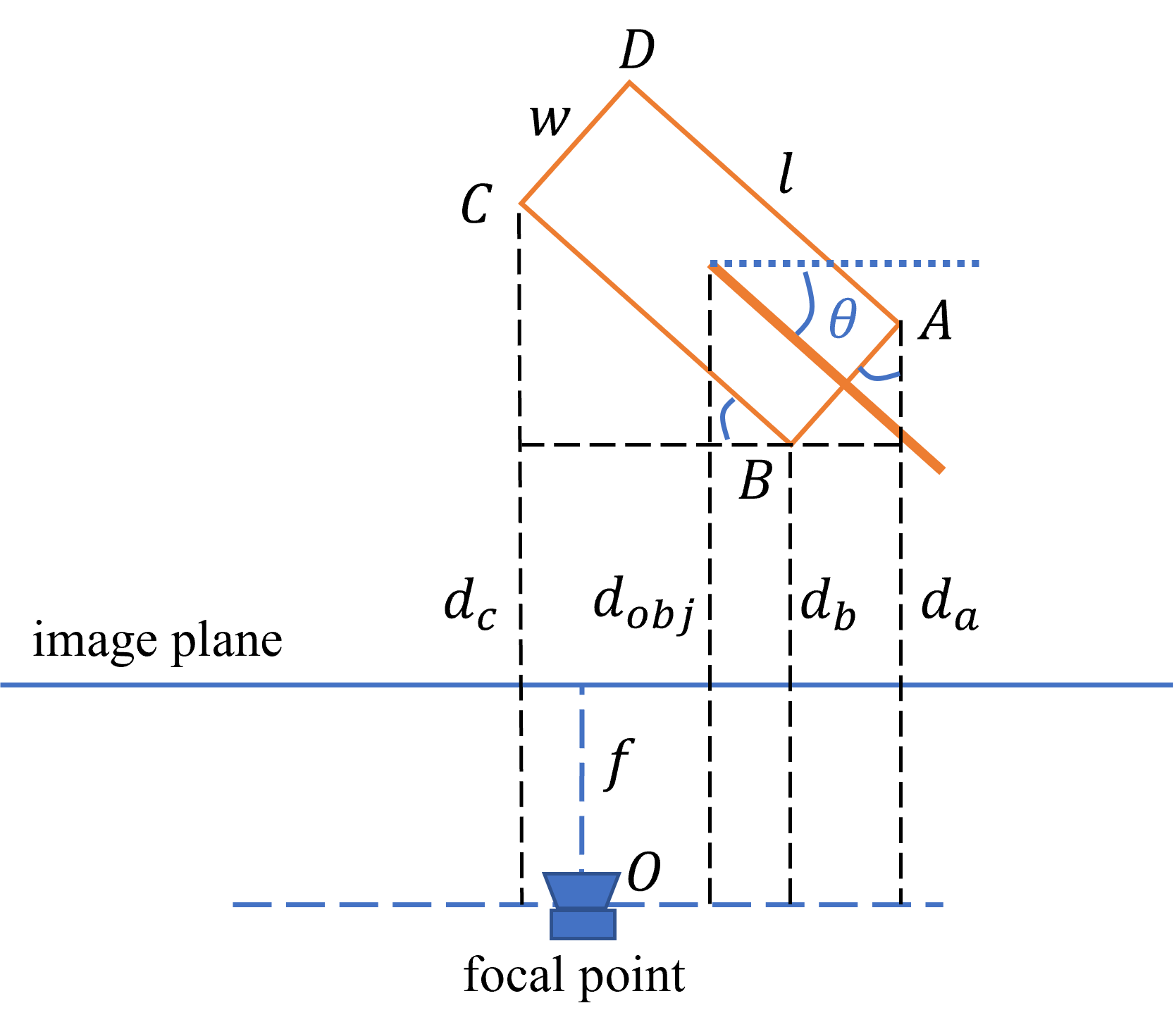}

  \caption{The geometry for calculating depth $d_{obj}$ and yaw angle $\theta$ from keyedge-ratios, shown in top-down view. The three angles denoted with blue arcs are identical. }
  \label{fig:geometry_dep}
\end{figure}

In this part we show that the depth and global yaw angle can be calculated from the keyedge-ratios and the physical size of the object. We show the math in the minimal setup, where two keyedge-ratios $r_{ba}$ and $r_{bc}$ are given, and the real length, width, and height of the object are denoted $l, w, h$, respectively. We also denote the focal length as $f$ (see  \cref{fig:geometry_dep}).

For any keyedge $i$, 
\begin{equation}\label{eq:height}
    \frac{f}{d_i} = \frac{h_i}{h} \Rightarrow d_a h_a = d_b h_b = d_c h_c = fh
\end{equation}

Denote the physical depth of the three corners $A, B, C$ (corresponding to the keyedges $a, b, c$) as $d_a, d_b, d_c$, we also have:
\begin{equation}\label{eq:dab}
    d_a = d_b + w \cos{\theta}, d_c = d_b + l \sin{\theta}
\end{equation}
\begin{equation}\label{eq:rba}
    \Rightarrow r_{ba} = \frac{h_b}{h_a} = \frac{d_a}{d_b} = 1 + \frac{w \cos{\theta}}{d_b}, r_{bc}=1 + \frac{l \sin{\theta}}{d_b}
\end{equation}

From \cref{eq:rba}, we get $d_b$ by cancelling $\theta$ and get $\theta$ by cancelling $d_b$:
\begin{align}\label{eq:db}
    \Rightarrow d_b & = \frac{1}{\sqrt{\frac{(r_{ba} - 1)^2}{w^2}+ \frac{(r_{bc} - 1)^2}{l^2}}} \\ \label{eq:theta}
    \theta & = \text{arctan2} ~ ~ (w(r_{bc} - 1), l(r_{ba} - 1))
\end{align}

Denote the depth of the object center as $d_{obj}$, we have
\begin{equation}\label{eq:dobj}
    d_{obj} = d_b + \frac{1}{2}(l \sin{\theta} + w \cos{\theta})
\end{equation}

In this way, the object depth $d_{obj}$ and yaw angle $\theta$ are obtained. Notice that the focal length is eliminated, and the final results only depend on $r_{ba}, r_{bc}, l, \text{and } w$. Therefore this method is \textit{camera-intrinsics-free}. For other keyedges, the angles and signs in the equations will change accordingly, which we show in more detail in the appendix.

\subsection{Network design}\label{sec:network}

\begin{figure*}[t]
  \centering
   \includegraphics[width=\textwidth]{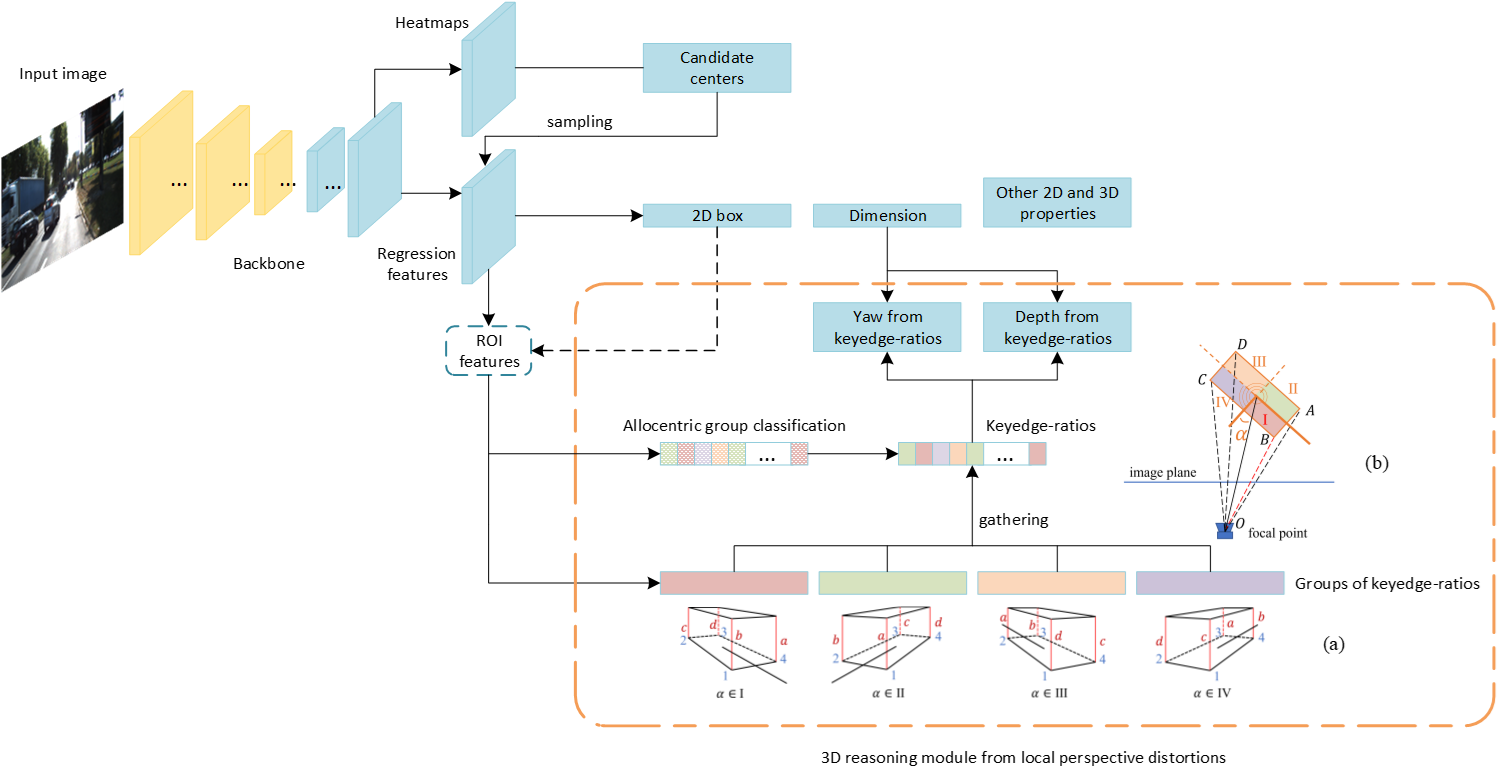}
   \caption{The overall structure of our network design. The orange dashed box highlights our local-perspective-based 3D inference module. Outside of the orange box is a typical one-stage or two-stage CNN architecture for Monocular 3D estimation, and we show it for the sake of completeness. The subfigure (a) illustrates the object-centric keyedge indices in red letters and the camera-centric keyedge indices in blue digits.  The allocentric groups are visualized in subfigure (b). 
   The dashed arrow and ROI-features are for two-stage networks only. }
   \label{fig:struct}
\end{figure*}

Overall, there are two main components in our module: keyedge-ratio regression with allocentric group classification and uncertainty-based depth fusion, as explained below. 

\subsubsection{Keyedge-ratio regression}\label{sec:kedge_idx}
The 3D inference based on local perspectives is independent of the absolute height of keyedges in an image. Therefore, we design a specific module to directly estimate keyedge-ratios.

First, we introduce a \textit{camera-centric indexing} strategy to regress keyedge-raitos. This strategy is illustrated in \cref{fig:struct} (a) as the blue numeric indices, in which index 1 is always assigned to the keyedge with shortest distance to the camera center. By using camera-centric indexing, all four keyedge-ratios $[r_{21}, r_{41}, r_{32}, r_{34}]$ are equal or smaller than 1 regardless of rotations (in most cases, see the appendix for further discussion), thus facilitating learning and convergence. 
The conversion between the camera-centric indexing $\{1,2,3,4\}$ and the object-centric indexing $\{a,b,c,d\}$ is through the \textit{allocentric group}, which is defined by the quarter that the allocentric angle $\alpha$ falls in, as visualized in \cref{fig:struct}~(b). Each allocentric group corresponds to a conversion between the camera-centric indices and object-centric indices as shown in \cref{fig:struct}~(b).

Second, we propose to use \textit{grouped heads} for the keyedge-ratio regression. In other words, different channels are used to regress keyedge-ratios for each allocentric group. In the network, there are 16 channels in total for the regression of 4 keyedge-ratios of all possible 4 allocentric groups. For each object, only the 4 channels corresponding to its allocentric group are used. The motivation is that the appearance of an object is very different when they are presented in different allocentric groups, and we decouple the learning for each of them to generate more discriminative features. 
As the keyedge-ratio regression relies on the allocentric group of an object, we also add the the 4-way allocentric group classification task to the network, supervised with cross-entropy loss.

\begin{table*}[t]
\centering
\resizebox{\textwidth}{!}{\begin{tabular}{l|ccc|ccc|ccc|ccc}
\toprule
\multirow{2}{*}{Methods} & \multicolumn{3}{c|}{Val, $AP_{3D}, IoU \geq 0.7$} & \multicolumn{3}{c|}{Val, $AP_{BEV}, IoU \geq 0.7$} & \multicolumn{3}{c|}{Test, $AP_{3D}, IoU \geq 0.7$} & \multicolumn{3}{c}{Test, $AP_{BEV}, IoU \geq 0.7$} \\ 
             & Easy           & Moderate          & Hard           & Easy            & Moderate          & Hard           & Easy            & Moderate          & Hard           & Easy            & Moderate           & Hard           \\ \midrule
MonoDIS \cite{simonelli2019disentangling}      & 11.06          & 7.60               & 6.37           & 18.45           & 12.58             & 10.66          & 10.37           & 7.94              & 6.40            & 17.23           & 13.19              & 11.12          \\
M3D-RPN \cite{brazil2019m3d}      & 14.53          & 11.07             & 8.65           & 20.85           & 15.62             & 11.88          & 14.76           & 9.71              & 7.42           & 21.02           & 13.67              & 10.23          \\
SMOKE \cite{liu2020smoke}       &       -         &          -         &       -        &        -         &        -           &       -         & 14.03           & 9.76              & 7.84           & 20.83           & 14.49              & 12.75          \\
MonoPair \cite{chen2020monopair}     & 16.28          & 12.30              & 10.42          & 24.12           & 18.17             & 15.76          & 13.04           & 9.99              & 8.65           & 19.28           & 14.83              & 12.89          \\
RTM3D \cite{li2020rtm3d}       &        -        &          -         &        -        &       -          &         -          &         -       & 14.41           & 10.34             & 8.77           & 19.17           & 14.20               & 11.99          \\
Movi3D \cite{simonelli2020towards}       & 14.28          & 11.13             & 9.68           & 22.36           & 17.87             & 15.73          & 15.19           & 10.90              & 9.26           & 22.76           & 17.03              & 14.85          \\
D4LCN \cite{ding2020learning}       & 22.32          & 16.20              & 12.30           & 31.53           & 22.58             & 17.87          & 16.65           & 11.72             & 9.51           & 22.51           & 16.02              & 12.55          \\
PGD \cite{wang2021probabilistic}         & \underline{24.35}          & \underline{18.34}             & \textbf{16.90}           & 30.56           & \underline{23.67}             & \underline{20.84}          & 19.05           & 11.76             & 9.39           & 26.89           & 16.51              & 13.49          \\
DDMP-3D \cite{wang2021depth}     &      -          &         -          &                &        -         &        -           &         -       & 19.71           & 12.78             & 9.80            & 28.08           & 17.89              & 13.44          \\
Ground-Aware \cite{liu2021ground} & 23.63          & 16.16             & 12.06          &        -         &           -        &      -          & \underline{21.65}           & 13.25             & 9.91           & \underline{29.81}           & 17.98              & 13.08          \\
CaDDN \cite{reading2021categorical}        & 23.57          & 16.31             & 13.84          &        -        &          -         &        -        & 19.17           & 13.41             & 11.46          & 27.94           & 18.91              & 17.19          \\
MonoGeo \cite{zhang2021learning}      & 18.45          & 14.48             & 12.87          & 27.15           & 21.17             & 18.35          & 18.85           & 13.81             & 11.52          & 25.86           & 18.99              & 16.19          \\
MonoEF \cite{zhou2021monocular}      &       -         &          -         &      -          &       -          &         -          &        -        & 21.29           & 13.87             & 11.71          & 29.03           & 19.70               & \underline{17.26}          \\
GUP Net \cite{lu2021geometry}     & 22.76          & 16.46             & 13.72          & 31.07           & 22.94              & 19.75           & 20.11           & 14.20               & 11.77           &      -            &       -             &      -           \\
AutoShape \cite{liu2021autoshape}    & 20.09          & 14.65             & 12.07          &        -         &             -       &       -          & \textbf{22.47}           & 14.17              & 11.36           & \textbf{30.66}            & \underline{20.08}              & 15.59           \\ \midrule
MonoRCNN \cite{shi2021geometry}    & 16.61          & 13.19             & 10.65          & 25.29           & 19.22             & 15.30           & 18.36           & 12.65             & 10.03          & 25.48           & 18.11              & 14.10           \\
\textit{Ours} (based on MonoRCNN)        & 18.44          & 14.60             & 12.57          & 26.19           & 20.67              & 17.30           & 19.74           & \underline{14.35}              & 11.94           & 27.52            & 20.07              & 16.34           \\ 
\textit{Improvement}  & \textit{+1.83}          & \textit{+1.41}             & \textit{+1.92}          & \textit{+0.90}           & \textit{+1.45}              & \textit{+2.00}           & \textit{+1.38}           & \textit{+1.70}              & \textit{+1.91}           & \textit{+2.04}            & \textit{+1.96}              & \textit{+2.24}           \\ \midrule
MonoFlex \cite{zhang2021objects}   & 23.64          & 17.51             & 14.83          & \underline{31.65}           & 23.29             & 20.02          & 19.94           & 13.89             & \underline{12.07}          & 28.23           & 19.75              & 16.89          \\
\textit{Ours} (based on MonoFlex)       & \textbf{25.66}  & \textbf{18.89}  & \underline{16.10}      & \textbf{33.71}   & \textbf{25.35}   & \textbf{22.18}   & 21.08            & \textbf{14.47}   & \textbf{12.73}   & 28.80              & \textbf{20.35}   & \textbf{17.57}   \\
\textit{Improvement} & \textit{+2.02}  & \textit{+1.38}  & \textit{+1.27}  & \textit{+2.06}   & \textit{+2.06}   & \textit{+2.16}   & \textit{+1.14}   & \textit{+0.58}   & \textit{+0.66}   & \textit{+0.57}    & \textit{+0.60}   & \textit{+0.68}  \\
\bottomrule
\end{tabular}%
}
\caption{Quantitative result on the KITTI validation set and test set. The best are highlighted in \textbf{bold} font. The second best are \underline{underlined}. The improvement over baselines are in \textit{italic} font. }
\label{tab:quan}
\end{table*}

\subsubsection{Uncertainty-based depth fusion}\label{sec:uncer}

After the conversion from camera-centric indexing to object-centric indexing, the keyedge-ratios are reorganized to 4 tuples: $(r_{ad}, r_{ab}), (r_{ba}, r_{bc}), (r_{cb}, r_{cd}), (r_{dc}, r_{da})$. Each tuple generates an estimation of depth and yaw angle as shown in \cref{sec:kedge_depth}, and we design a dedicated uncertainty based fusion module to generate the final result.


Specifically, each keyedge-ratio prediction head $r_{ij}$ is accompanied with a regressed uncertainty estimation value $\sigma_{ij}$. Denote the depth estimated from keyedge ratios as $d_{kr}$, we approximate the uncertainty of $d_{kr}$ using the first order derivative: 
\begin{equation}
    \sigma_{d_{kr}} = \sum_{(i,j)} \frac{\partial d_{kr}}{\partial r_{ij}}\sigma_{ij}
\end{equation}
The partial derivatives $\frac{\partial d_{kr}}{\partial r_{ij}}$ are calculated using PyTorch's automatic differentiation engine \cite{NEURIPS2019_9015}. 

The depth estimated from the keyedge-ratio tuples are averaged using the inverse uncertainty as the weight, $w_{d_t} = 1 / \sigma_{d_t}, d_{fusion} = \sum_i d_t w_{d_t}$, where $t$ indexes each depth prediction. The uncertainty-based weighted sum is also used when combining this method with other depth predictions in an existing network, as in \cref{sec:exp}. 
The keyedge-ratios and their uncertainties are together supervised by an uncertainty-aware loss function \cite{10.5555/3295222.3295309}:
\begin{equation}
    L(r_{ij}, \sigma_{ij}; r_{ij}^*) = \frac{| r_{ij} - r_{ij}^* |}{\sigma_{ij}} + \log \sigma_{ij}
\end{equation}
where $r_{ij}^*$ is the ground truth keyedge ratio.

\subsubsection{Components outside of the local perspective module}
Aside from the proposed local perspective module estimating the keyedge-ratios as introduced above, there are other variables to be estimated to complete the monocular 3D object detection task. As our local perspective module can be plugged with various network structures, the list of variables estimated outside of the module slightly varies depending on the overall network structure used. Specifically, we incorporate our local perspective module with three networks: MonoFlex~\cite{zhang2021objects}, MonoRCNN~\cite{shi2021geometry}, and PGD~\cite{wang2021probabilistic}, as they are representative one-stage and two-stage networks on KITTI~\cite{Geiger2012CVPR} and nuScenes~\cite{nuscenes2019} benchmarks. 

Commonly estimated variables across the networks include object classification score, 2D projected center, and physical size of 3D bounding box. The estimated physical length $l$ and width $w$ are used in \cref{eq:db,eq:theta,eq:dobj}. The 2D bounding box and keypoints (defined by the 2D projection of 3D bounding box) are regressed in MonoFlex~\cite{zhang2021objects} and MonoRCNN \cite{shi2021geometry}, but are optional in PGD~\cite{wang2021probabilistic}. The list of variables and their corresponding regression heads and loss functions follow the original networks. 

\cref{fig:struct} is an overall illustration of the common architecture of the networks we incorporated with in this work. The dashed orange rectangle is the proposed local-perspective module, and outside of it are the backbone and regression heads of the plugged-in networks. The dashed arrow and the ROI-feature block are for MonoRCNN~\cite{shi2021geometry} as a two-stage network in which the regression heads are after extracting ROI features, while the one-stage networks (MonoFlex~\cite{zhang2021objects} and PGD~\cite{wang2021probabilistic}) follow the solid arrows.

\section{Experiments}\label{sec:exp}
The experiments are conducted on the KITTI dataset \cite{Geiger2012CVPR} and nuScenes dataset\cite{nuscenes2019}. KITTI has 7,481 images for training and 7,518 images for testing. We further split the official training set into training and validation set following \cite{chen20153d} for fair comparison with other baselines. The dataset evaluates three object classes: Car, Pedestrian, and Cyclist, and we mainly focus on the Car category when reporting the results following previous works. The evaluation metric is Average Precision of the 3D bounding boxes ($AP_{3D}$) and of the bird's eye-view 2D rotated bounding boxes ($AP_{BEV}$). We use 40 recall positions, which is the more meaningful metric compared with the 11-recall-position version, according to \cite{simonelli2019disentangling}. We also extend our experiments to the nuScenes dataset \cite{nuscenes2019} to show that our method works under various environments. 
\subsection{Experiments on KITTI dataset}

\subsubsection{Implementation details}
For the experiment on KITTI dataset, we choose two baselines, MonoFlex \cite{zhang2021objects} and MonoRCNN \cite{shi2021geometry}, and implement our method based on these networks. They are representative one-stage and two-stage networks for monocular 3D object detection respectively. We fuse the local-perspective- (LP-) based estimation with the baselines' original estimation to show that our method brings value beyond existing approaches and can be incorporated to improve them. 
Both baselines have uncertainty estimation for their output, facilitating the uncertainty-based fusion as explained in \cref{sec:uncer}. The training setups (batch size, training epochs, optimizer settings, data augmentations) are consistent with the baselines. 

\begin{table*}[t]
\centering
\resizebox{\linewidth}{!}{
\begin{tabular}{cc|ccc|ccc|ccc|ccc}
\toprule
\multicolumn{2}{c|}{Baselines} & \multicolumn{6}{c|}{MonoFlex}                                                                               & \multicolumn{6}{c}{MonoRCNN}                                                                               \\ \midrule
                    \multicolumn{2}{c|}{Methods}   & \multicolumn{3}{c|}{Val, $AP_{3D}, IoU \geq 0.7$} & \multicolumn{3}{c|}{Val, $AP_{BEV}, IoU \geq 0.7$} & \multicolumn{3}{c|}{Val, $AP_{3D}, IoU \geq 0.7$} & \multicolumn{3}{c}{Val, $AP_{BEV}, IoU \geq 0.7$} \\
                 B        & LP       & Easy           & Moderate          & Hard           & Easy            & Moderate          & Hard           & Easy           & Moderate          & Hard           & Easy            & Moderate          & Hard           \\ \midrule
\checkmark        &          & 23.64          & 17.51             & 14.83          & 31.65           & 23.29             & 20.02          & 16.61          & 13.19             & 10.65          & 25.29           & 19.22             & 15.30          \\
          & \checkmark        & 21.13          & 16.08             & 14.07          & 28.91           & 21.29             & 18.91          &11.98          & 9.81              & 8.41           & 17.61           & 14.14             & 12.49 \\
  \checkmark        & \checkmark        & \textbf{25.66}          & \textbf{18.89}             & \textbf{16.10}          & \textbf{33.71}           & \textbf{25.35}             & \textbf{22.18}          & \textbf{18.44}          & \textbf{14.60}             & \textbf{12.57}          & \textbf{26.19}           & \textbf{20.67}             & \textbf{17.30}          \\ 
 \bottomrule
\end{tabular}
}
\caption{Comparison of different settings on KITTI validation set. \textbf{B} represents the estimation method in the baseline. \textbf{LP} represents our local-perspective estimation with keyedge-ratio regression. }
\label{tab:roirot}
\end{table*}

\begin{table*}[t]
\centering
\resizebox{\textwidth}{!}{
\begin{tabular}{l|cccccc|cccccc}
\toprule
Methods                 & \multicolumn{6}{c|}{Val, $AP_{3D}, IoU \geq 0.7$}                                & \multicolumn{6}{c}{Val, $AP_{BEV}, IoU \geq 0.7$}                               \\
                        & \multicolumn{2}{c}{Easy} & \multicolumn{2}{c}{Moderate} & \multicolumn{2}{c|}{Hard} & \multicolumn{2}{c}{Easy} & \multicolumn{2}{c}{Moderate} & \multicolumn{2}{c}{Hard} \\ \midrule
Ours (based on MonoFlex)           & 25.66  &                 & 18.89    &                   & 16.10   &                 & 33.71  &                 & 25.35    &                   & 22.18  &                 \\ \midrule
no grouped heads                            & 23.01  & \textit{-2.65}  & 17.6     & \textit{-1.29}    & 14.82  & \textit{-1.28}  & 31.26  & \textit{-2.45}  & 23.48    & \textit{-1.87}    & 20.02  & \textit{-2.16}  \\
no camera-centric indexing                  & 22.81  & \textit{-2.85}  & 16.31    & \textit{-2.58}    & 13.71  & \textit{-2.39}  & 30.44  & \textit{-3.27}  & 22.35    & \textit{-3.00}       & 19.45  & \textit{-2.73}  \\
no grouped heads \& camera-centric indexing & 21.21  & \textit{-4.45}  & 15.44    & \textit{-3.45}    & 13.20   & \textit{-2.90}   & 29.66  & \textit{-4.05}  & 21.30     & \textit{-4.05}    & 18.37  & \textit{-3.81} \\
\bottomrule
\end{tabular}
}
\caption{Ablation study on the KITTI validation set. Performance change compared with our default setup is highlighted in \textit{italic} font. }
\label{tab:ablation}
\end{table*}

\subsubsection{Quantitative and qualitative results}
\begin{figure*}[t]
  \centering
   \includegraphics[width=\linewidth]{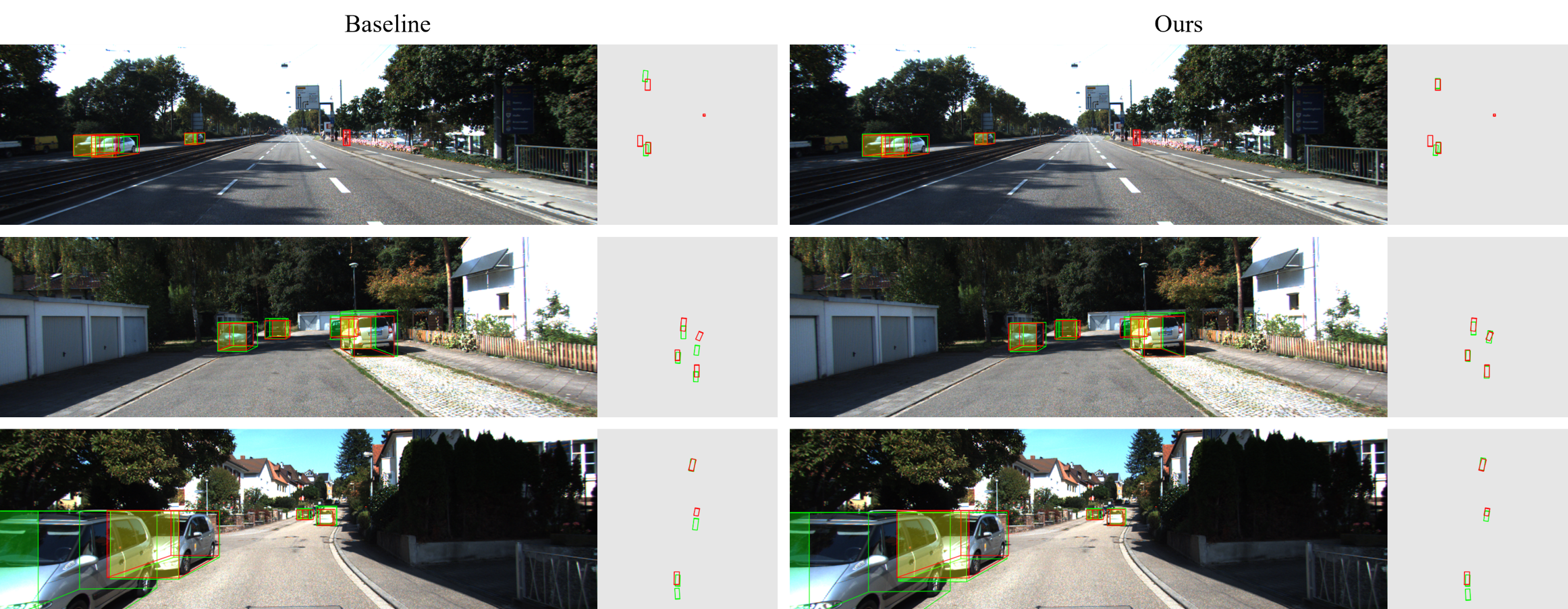}
   \caption{Qualitative results on the KITTI validation set. In each column, the image is on the left, and 3D detection results in bird's-eye-view are shown on the right. The ground truths are in red, and the detections are in green. We achieve general improvements over the baseline.}
   \label{fig:qual}
\end{figure*}

The quantitative results are shown in \cref{tab:quan}. Our method achieves consistent improvements on all evaluated metrics in the validation set and the test set over both baselines. The number of parameters in the network without and with our LP module is 21.47M v.s. 22.07M (for MonoFlex) and 69.71M v.s. 70.80M (for MonoRCNN), accounting for only 2.80\% and 1.56\% increase respectively. Therefore we account the improvement to the proposed methodology instead of the added parameters. 
The inference time overhead of our LP module is also limited, accounting for less than 10\% increase (0.034s v.s. 0.037s for MonoFlex and 0.06s v.s. 0.065s for MonoRCNN). 
Qualitative examples of our results (with MonoFlex baseline) are shown in \cref{fig:qual}. Our method delivers more accurate localization performance. 

\subsubsection{Effect of local perspectives in estimation}
In \cref{tab:roirot} we further decouple the effect of local perspective estimation. As shown in the second row, using keyedge-ratio regression alone for 3D detection yields suboptimal results. It is not surprising since the local perspective distortion of objects are more subtle to estimate than their size and position in an image. It is not a problem since we do not need to discard the stronger signal and restrict ourselves to local perspectives alone in practice. Nonetheless, the local-perspective-based approach shows its value when combined with the existing approaches. The universal improvements in all metrics for both baselines indicate that the local perspective distortion is a missing piece in previous work, and we may incorporate it with various existing approaches for general improvements. 




\begin{figure}[t]
  \centering
   \includegraphics[width=0.9\linewidth]{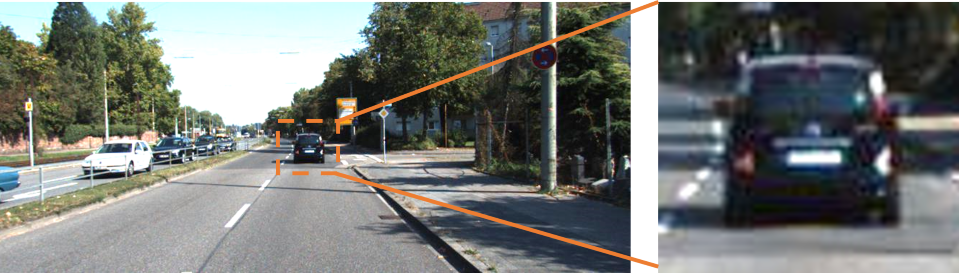}
   \caption{Example of a vehicle in the up front showing little local perspective distortion. In \cref{sec:upfront} we show that the performance of LP module does not deteriorate on such objects. }
   \label{fig:upfront}
\end{figure}

\begin{figure}[t]
  \centering
   \includegraphics[width=0.9\linewidth]{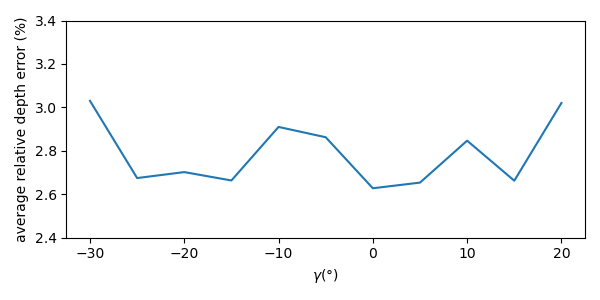}
   \caption{Analysis of the error of local perspective estimation for objects at different viewpoints, evaluated on car category of KITTI validation set. }
   \label{fig:gamma}
\end{figure}

\subsubsection{Ablation study on Keyedge-ratio regression}
Here we validate that the specific design of the keyedge-ratio regression head is beneficial. We experimented on the MonoFlex-based network, removing the grouped heads by using the same four channels to regress the four keyedge-ratios regardless of the allocentric group. We also experimented using the object-centric indexing instead of the camera-centric indexing. 

\begin{table*}[t]
\centering
\resizebox{\textwidth}{!}{
\begin{tabular}{l|c|cccccccccc}
\toprule
Methods                  & mAP   & car   & truck & bus   & trailer & \begin{tabular}[c]{@{}c@{}}construction\\ \_vehicle\end{tabular} & pedestrian & motorcycle & bicycle & \begin{tabular}[c]{@{}c@{}}traffic\\ \_cone\end{tabular} & barrier \\ \midrule
MonoFlex*                 & 0.167 & 0.336 & 0.080 & 0.177 & 0.023   & 0.000                 & 0.238     & 0.104      & 0.057   & 0.330         & 0.325   \\
\textit{Ours} (based on MonoFlex)* & 0.182 & 0.353 & 0.099 & 0.175 & 0.016   & 0.004                 & 0.255     & 0.152      & 0.081   & 0.347         & 0.337  \\ \midrule
PGD                 & 0.317 & 0.502 & 0.226 & 0.291 & 0.074   & 0.043                 & 0.425     & 0.312      & 0.287   & 0.551         & 0.462   \\
\textit{Ours} (based on PGD) & 0.321 & 0.512 & 0.233 & 0.294 & 0.070   & 0.044                 & 0.427     & 0.317      & 0.295   & 0.555         & 0.460  \\ \bottomrule
\end{tabular}


}
\caption{Experiment results on nuScenes validation set. The AP definition is different from that in KITTI. * means that the results are only on images from the FRONT camera. }
\label{tab:nusc}
\end{table*}


The experimental results are shown in \cref{tab:ablation}. The removal of either the grouped heads or the camera-centric indexing causes severe deterioration (up to $3$ points), and the results are the worst when both are absent. It demonstrates that the proposed regression head design is crucial to the accurate estimation of keyedge-ratios.


\subsubsection{Observability of local perspectives}\label{sec:upfront}
One may suspect that the local perspective distortion is only observable for objects with certain viewpoints, which may limit the general applicability of the proposed approach. For example, a vehicle in front of the camera with only the rear face visible shows very little clue about its local perspective distortion. An example is shown in \cref{fig:upfront}. Does our method fail on such objects? The answer is no, as shown in the following.

To inspect this issue, we analyze the depth estimation accuracy for objects at different viewpoints, parameterized by the viewing angle $\gamma$ as defined in \cref{sec:geometry}. The underlying assumption is that the majority of vehicles in the up front are also axis-aligned, showing only a single face to the camera. It is generally true for the KITTI dataset. If the local perspective does not work for objects up-front, the depth estimation would deteriorate near $\gamma=0$. Since objects in the center of view (with $\gamma$ close to zero) tend to have larger depth, the error of estimated depth also tends to be larger. We rule out the effect of different distribution of ground truth depth at different $\gamma$ by evaluating the relative depth error, i.e., $d_r = | d_{est} - d_{gt} | / d_{gt}$. However, it is not straight-forward to evaluate the (relative) depth error in a 3D detection task, because there is no clear definition on pairs of corresponding ground truth objects and estimations, which depends on detection association and confidence thresholds. Here we mimic \textit{AOS} (average orientation similarity) in KITTI \cite{Geiger2012CVPR} and design a new metric called average relative depth error (\textit{ARDE}):
\begin{equation}
    ARDE = \frac{1}{40} \sum_{r\in [r_{40}]} \max_{\Tilde{r}\geq r} s(\Tilde{r})
\end{equation}
 where
\begin{equation}
    s(r) = \frac{1}{|\mathcal{D}(r)|} \sum_{i\in \mathcal{D}(r)} d_r(i)
\end{equation}
in which $[r_{40}]$ is the set of 40 recall-rate points used in calculating average precision (\textit{AP}), $\mathcal{D}(r)$ is the set of true positive detections at recall rate $r$, and $d_r(i)$ is the relative depth error of detection $i$. Here the \textit{recalls} are w.r.t. the 2D bounding box detections, consistent with \textit{AOS} in KITTI. This metric reflects the relative depth error for objects with various detection confidence. In \cref{fig:gamma}, we show the result of depth error from local perspectives (LP) only based on the MonoFlex baseline. The range of $\gamma$ is set to $[-30^{\circ}, 20^{\circ}]$ because objects outside of the range are of far less frequency. It turns out that there is no obvious pattern in the \textit{ARDE} w.r.t. $\gamma$ the viewing angle, indicating that the LP module does not suffer from up-front objects. The reason might be that the network is able to infer keyedge-ratios from the global information, exploiting the connection between global perspective information (position, size, etc.) and the local perspective distortion implicitly. 



\subsection{Experiments on nuScenes dataset}
We conduct further experiments on the nuScenes \cite{nuscenes2019} dataset to show that the value of our method is generally valid across different environments. 

First we retrain the MonoFlex\cite{zhang2021objects}-based networks without finetuning hyperparameters for nuScenes. Since MonoFlex only conducted experiments on KITTI, the baseline results are generated by us, and we only work on the images from the FRONT camera which have similar viewpoints as KITTI, resulting in 28,130/6,019 images in the training/validation set. We train 100 epochs with batch size 16 on 8 GPUs. 

Then we also incorporate the local perspective module with PGD~\cite{wang2021probabilistic}, which reported results on the nuScenes dataset. The official open-sourced implementation of PGD does not include the geometric graph part, thus the baseline used in the following comparison and the network fused with our local perspective module both do not have the geometric graph part. We follow the official split of nuScenes, resulting in 106,416 /36,114 images in the training and validation set. We follow the "1x" schedule of PGD, i.e., training 12 epochs with batch size 16 on 8 GPUs. As only the learning schedules for evaluation on the validation set are released by PGD, our result is also on the validation set. 


The experiment result is in \cref{tab:nusc}. 
Here we report the Average Prevision (AP) metric defined by the official nuScenes benchmark, measured with the 2D center distance on the ground plane and averaged over thresholds of 0.5, 1, 2, 4 meters. mAP is calculated by taking the mean over all classes. 
We achieved higher mAP and higher AP in most categories, compared with both baselines. 

\section{Conclusion}
We propose a method to leverage the local perspective effect for monocular 3D object detection. We design a new regression target called keyedge ratios to parameterize the local perspective distortion. Given a pair of keyedge ratios and the physical size of an object, the proposed method estimates the depth and the yaw angle of objects without camera intrinsic or extrinsic parameters. 

The task of keyedge ratio regression is not restrictive to a specific network architecture, and can be appended to arbitrary backbones and middle layers. Therefore this work can be regarded as a generic plug-in module which may be used to augment a wide series of existing work. We incorporate the local-perspective-based module with three recent representative monocular 3D object detection networks and achieve consistent improvements on the KITTI dataset and the nuScenes dataset. 

Our work also has some limitations. While the local-perspective-based method explores a new way for 3D reasoning, it needs to be combined with existing methods (e.g., based on visual size and position) to bring improvements. The method is not designed for objects that are very far away, in which case the local perspective distortion diminishes. As some works that exploit the global perspectives already emerge, it is an interesting direction to incorporate the global and local perspective effects together to improve the estimation of both of them. We believe our approach provides a novel view and additional opportunities for future research on monocular 3D object detection. 

{\small
\bibliographystyle{wacv2023/ieee_fullname}
\bibliography{wacv2023/mybib}
}

\end{document}


\title{MonoEdge: Monocular 3D Object Detection Using Local Perspectives (Appendix)}

\author{First Author\\
Institution1\\
Institution1 address\\
{\tt\small firstauthor@i1.org}
\and
Second Author\\
Institution2\\
First line of institution2 address\\
{\tt\small secondauthor@i2.org}
}

\maketitle
\thispagestyle{empty}

\section{Depth and yaw from keyedge-ratios}
\label{sec:deriv}
In the paper, we derived the depth $d_{obj}$ and yaw angle $\theta$ from a pair of keyedge-ratios $(r_{ba}, r_{bc})$. The calculation for other keyedge-ratios are similar. Given all four keyedge-ratios $r_{ab}, r_{bc}, r_{cd}, r_{da}$, we first reorganize them into four tuples $(r_{ad}, r_{ab}), (r_{ba}, r_{bc}), (r_{cb}, r_{cd}), (r_{dc}, r_{da})$. Each tuple has a reference keyedge which occurs in both keyedge-ratios (e.g., $a$ in $(r_{ad}, r_{ab})$). We use a common notation $(r_1, r_2)$ to represent the first and second elements of each tuple. Then for each tuple, the solution of depth and yaw has the same form:
\begin{equation}\label{eq:theta}
    \theta = \text{arctan2} ~ ~ (w R^\theta_w, l R^\theta_l)
\end{equation}
\begin{equation}\label{eq:di}
    d_i = \frac{1}{\sqrt{\frac{{R^d_w}^2}{w^2}+ \frac{{R^d_l}^2}{l^2}}}
\end{equation}
where $\theta$ is the yaw angle and $d_i$ is the depth of the reference keyedge. $R^\theta_w$, $R^\theta_l$, $R^d_w$, and $R^d_l$ are placeholders of which the value follows \cref{tab:placeholders}. 

Then the depth of the object center is:
\begin{equation}\label{eq:depth}
    d_{obj} = d_i + \frac{1}{2}\Delta_d
\end{equation}
where the value of $\Delta_d$ is listed in \cref{tab:depth}. 

As discussed in the paper, the depth and yaw derived from keyedge-ratios only depends on the pair of the keyedge-ratios and the physical length $l$ and width $w$ of the object. 

\section{Camera-centric indexing}
With camera-centric indexing, index 1 is always assigned to the keyedge with shortest distance to the camera center. Notice that the distance is not the depth (which is the projected distance in the front direction). When the keyedge with the shortest distance also has the smallest depth, all four keyedge-ratios $[r_{21}, r_{41}, r_{32}, r_{34}]$ are equal or smaller than 1. Otherwise, the keyedge-ratios can go slightly larger than 1. In practice, the keyedge with shortest depth has shortest distance for most objects in our tested datasets. We use the distance for camera-centric indexing because it is invariant to perspective rotations. 

\begin{table}[t]
\centering
\resizebox{0.9\linewidth}{!}{%
\begin{tabular}{c|cccc}
\toprule
\begin{tabular}[c]{@{}c@{}}reference\\  keyedge\end{tabular} & $R^\theta_w$ & $R^\theta_l$ & $R^d_w$ & $R^d_l$  \\ \hline
$a$                                                           &  $r_1 - 1$  & $-(r_2 - 1)$  & $r_2 - 1$ & $r_1 - 1$  \\
$b$                                                           &  $r_2 - 1$  &  $r_1 - 1$  & $r_1 - 1$ & $r_2 - 1$  \\
$c$                                                           &  $-(r_1 - 1)$  &  $r_2 - 1$  & $r_2 - 1$ & $r_1 - 1$  \\
$d$                                                           & $-(r_2 - 1)$   &  $-(r_1 - 1)$  & $r_1 - 1$ & $r_2 - 1$  \\ \bottomrule
\end{tabular}%
}
\caption{Value of the placeholders in \cref{eq:theta} and \cref{eq:di} for each tuple of keyedge-ratios. }
\label{tab:placeholders}
\end{table}



\begin{table}[t]
\centering
\resizebox{0.5\linewidth}{!}{%
\begin{tabular}{c|c}
\toprule
\begin{tabular}[c]{@{}c@{}}reference\\  keyedge\end{tabular} & $\Delta_d$  \\ \hline
$a$                                                           &  $l\sin \theta - w \cos \theta$ \\
$b$                                                           &  $l\sin \theta + w \cos \theta$ \\
$c$                                                           &  $-l \sin \theta + w \cos \theta$ \\
$d$                                                           &  $ -l \sin \theta - w \cos \theta$ \\ \bottomrule
\end{tabular}%
}
\caption{Value of $\Delta_d$ in \cref{eq:depth} for each tuple of keyedge-ratios. }
\label{tab:depth}
\end{table}



\title{MonoEdge: Monocular 3D Object Detection Using Local Perspectives (Appendix)}

\author{Minghan Zhu\textsuperscript{1*}, 
Lingting Ge\textsuperscript{2}, 
Panqu Wang\textsuperscript{2}, 
Huei Peng\textsuperscript{1}\\
\textsuperscript{1}University of Michigan\\
\textsuperscript{2}TuSimple Inc\\
{\tt\small minghanz@umich.edu, lingting.ge@tusimple.ai, panqu.wang@tusimple.ai, hpeng@umich.edu}
\thanks{*Work done during internship at TuSimple Inc. }
}

\maketitle
\thispagestyle{empty}

\section{Depth and yaw from keyedge-ratios}
\label{sec:deriv}
In the paper, we derived the depth $d_{obj}$ and yaw angle $\theta$ from a pair of keyedge-ratios $(r_{ba}, r_{bc})$. The calculation for other keyedge-ratios are similar. Given all four keyedge-ratios $r_{ab}, r_{bc}, r_{cd}, r_{da}$, we first reorganize them into four tuples $(r_{ad}, r_{ab}), (r_{ba}, r_{bc}), (r_{cb}, r_{cd}), (r_{dc}, r_{da})$. Each tuple has a reference keyedge which occurs in both keyedge-ratios (e.g., $a$ in $(r_{ad}, r_{ab})$). We use a common notation $(r_1, r_2)$ to represent the first and second elements of each tuple. Then for each tuple, the solution of depth and yaw has the same form:
\begin{equation}\label{eq:theta}
    \theta = \text{arctan2} ~ ~ (w R^\theta_w, l R^\theta_l)
\end{equation}
\begin{equation}\label{eq:di}
    d_i = \frac{1}{\sqrt{\frac{{R^d_w}^2}{w^2}+ \frac{{R^d_l}^2}{l^2}}}
\end{equation}
where $\theta$ is the yaw angle and $d_i$ is the depth of the reference keyedge. $R^\theta_w$, $R^\theta_l$, $R^d_w$, and $R^d_l$ are placeholders of which the value follows \cref{tab:placeholders}. 

Then the depth of the object center is:
\begin{equation}\label{eq:depth}
    d_{obj} = d_i + \frac{1}{2}\Delta_d
\end{equation}
where the value of $\Delta_d$ is listed in \cref{tab:depth}. 

As discussed in the paper, the depth and yaw derived from keyedge-ratios only depends on the pair of the keyedge-ratios and the physical length $l$ and width $w$ of the object. 

\section{Camera-centric indexing}
With camera-centric indexing, index 1 is always assigned to the keyedge with shortest distance to the camera center. Notice that the distance is not the depth (which is the projected distance in the front direction). When the keyedge with the shortest distance also has the smallest depth, all four keyedge-ratios $[r_{21}, r_{41}, r_{32}, r_{34}]$ are equal or smaller than 1. Otherwise, the keyedge-ratios can go slightly larger than 1. In practice, the keyedge with shortest depth has shortest distance for most objects in our tested datasets. We use the distance for camera-centric indexing because it changes accordingly with the allocentric angle and visible faces of an object, which largely affects the appearance of the object in an image. 

\begin{table}[t]
\centering
\resizebox{0.9\linewidth}{!}{%
\begin{tabular}{c|cccc}
\toprule
\begin{tabular}[c]{@{}c@{}}reference\\  keyedge\end{tabular} & $R^\theta_w$ & $R^\theta_l$ & $R^d_w$ & $R^d_l$  \\ \hline
$a$                                                           &  $r_1 - 1$  & $-(r_2 - 1)$  & $r_2 - 1$ & $r_1 - 1$  \\
$b$                                                           &  $r_2 - 1$  &  $r_1 - 1$  & $r_1 - 1$ & $r_2 - 1$  \\
$c$                                                           &  $-(r_1 - 1)$  &  $r_2 - 1$  & $r_2 - 1$ & $r_1 - 1$  \\
$d$                                                           & $-(r_2 - 1)$   &  $-(r_1 - 1)$  & $r_1 - 1$ & $r_2 - 1$  \\ \bottomrule
\end{tabular}%
}
\caption{Value of the placeholders in \cref{eq:theta} and \cref{eq:di} for each tuple of keyedge-ratios. }
\label{tab:placeholders}
\end{table}



\begin{table}[t]
\centering
\resizebox{0.5\linewidth}{!}{%
\begin{tabular}{c|c}
\toprule
\begin{tabular}[c]{@{}c@{}}reference\\  keyedge\end{tabular} & $\Delta_d$  \\ \hline
$a$                                                           &  $l\sin \theta - w \cos \theta$ \\
$b$                                                           &  $l\sin \theta + w \cos \theta$ \\
$c$                                                           &  $-l \sin \theta + w \cos \theta$ \\
$d$                                                           &  $ -l \sin \theta - w \cos \theta$ \\ \bottomrule
\end{tabular}%
}
\caption{Value of $\Delta_d$ in \cref{eq:depth} for each tuple of keyedge-ratios. }
\label{tab:depth}
\end{table}

\section{More details on the experimented networks}
Given that our proposed local perspective module can be plugged-in to various network structures, here we give more details on the regressed variables and corresponding loss functions on the experimented networks outside of the local perspective module, so that readers have a better idea of the overall architecture. 
\subsection{MonoFlex~\cite{zhang2021objects}}
The regressed variables are:
\begin{itemize}
    \item Classification score (focal loss~\cite{lin2017focal})%
    \item 2D bounding box (GIoU loss~\cite{rezatofighi2019generalized})%
    \item 2D projected center (L1 loss)
    \item Projected keypoints (L1 loss)
    \item Physical size (L1 loss)
    \item Depth and its uncertainty (Uncertainty-aware loss, same form as Eq. (9))
\end{itemize}
The size of the regression head of the local perspective module follows the design of other regression heads of this network, i.e., 1 FC layer with 256 dimensional features. 

\subsection{MonoRCNN~\cite{shi2021geometry}}
The regressed variables are:
\begin{itemize}
    \item Classification score (cross entropy loss)
    \item 2D bounding box (L1 loss)
    \item 2D projected center (L1 loss)
    \item Projected keypoints (L1 loss)
    \item Physical size (L1 loss)
    \item Physical height and inverse visual height and their uncertainty for depth estimation (Uncertainty-aware loss, same form as Eq. (9))
\end{itemize}
The size of the regression head of the local perspective module follows the design of other regression heads of this network, i.e., 2 FC layer with 1024 dimensional features. 

\subsection{PGD~\cite{wang2021probabilistic}}
The regressed variables are:
\begin{itemize}
    \item Classification score (cross entropy loss)
    \item Centerness (BCE loss)
    \item 2D projected center (smooth L1 loss)
    \item 2D bounding box (smooth L1 loss)
    \item Physical size (smooth L1 loss)
    \item Velocity (smooth L1 loss)
    \item Depth and its weight (smooth L1 loss)
\end{itemize}
The size of the regression head of the local perspective module follows the design of other regression heads of this network, i.e., 1 FC layer with 256 dimensional features. 

{\small
\bibliographystyle{ieee_fullname}
\bibliography{mybib}
}